\documentclass[conference]{IEEEtran}
\IEEEoverridecommandlockouts

\usepackage{cite}
\usepackage{amsmath,amssymb,amsfonts}
\usepackage{algorithmic}
\usepackage{graphicx}
\usepackage{textcomp}

\usepackage{hyperref}       
\usepackage{url}            
\usepackage{booktabs}       
\usepackage{nicefrac}       
\usepackage{microtype}      

\usepackage{bbm}
\usepackage{array,multirow}
\usepackage{duckuments}
\usepackage[export]{adjustbox}
\usepackage{arydshln}
\usepackage{caption}
\usepackage{wrapfig}
\usepackage[ruled,vlined]{algorithm2e}
\usepackage[table]{xcolor}

\SetCommentSty{mycommfont}

\SetKwInput{KwInput}{Require}                
\SetKwInput{KwOutput}{Output}              
\expandafter\def\expandafter\normalsize\expandafter{%
    \normalsize%
    \setlength\abovedisplayskip{0pt}%
    \setlength\belowdisplayskip{-5pt}%
    \setlength\abovedisplayshortskip{-8pt}%
    \setlength\belowdisplayshortskip{-5pt}%
}

\def\BibTeX{{\rm B\kern-.05em{\sc i\kern-.025em b}\kern-.08em
    T\kern-.1667em\lower.7ex\hbox{E}\kern-.125emX}}

\makeatletter
\newcommand{\removelatexerror}{\let\@latex@error\@gobble}
\makeatother

\begin{document}

\title{BloomCoreset: Fast Coreset Sampling using Bloom Filters for Fine-Grained Self-Supervised Learning
}

\author{\IEEEauthorblockN{Prajwal Singh}
\IEEEauthorblockA{\textit{CVIG Lab} \\
\textit{IIT Gandhinagar}\\
Gandhinagar, India \\
singh\_prajwal@iitgn.ac.in}
\and
\IEEEauthorblockN{Gautam Vashishtha}
\IEEEauthorblockA{\textit{CVIG Lab} \\
\textit{IIT Gandhinagar}\\
Gandhinagar, India \\
gautam.pv@iitgn.ac.in}
\and
\IEEEauthorblockN{Indra Deep Mastan}
\IEEEauthorblockA{\textit{Computer Science \& Eng.} \\
\textit{IIT BHU}\\
Varanasi, India \\
indra.cse@itbhu.ac.in}
\and
\IEEEauthorblockN{Shanmuganathan Raman}
\IEEEauthorblockA{\textit{CVIG Lab} \\
\textit{IIT Gandhinagar}\\
Gandhinagar, India \\
shanmuga@iitgn.ac.in}
}

\maketitle


\begin{abstract}
The success of deep learning in supervised fine-grained recognition for domain-specific tasks relies heavily on expert annotations. The Open-Set for fine-grained Self-Supervised Learning (SSL) problem aims to enhance performance on downstream tasks by strategically sampling a subset of images (the Core-Set) from a large pool of unlabeled data (the Open-Set). In this paper, we propose a novel method, BloomCoreset, that significantly reduces sampling time from Open-Set while preserving the quality of samples in the coreset. To achieve this, we utilize Bloom filters as an innovative hashing mechanism to store both low- and high-level features of the fine-grained dataset, as captured by Open-CLIP, in a space-efficient manner that enables rapid retrieval of the coreset from the Open-Set. To show the effectiveness of the sampled coreset, we integrate the proposed method into the state-of-the-art fine-grained SSL framework, SimCore \cite{kim2023coreset}. The proposed algorithm drastically outperforms the sampling strategy of the baseline in \cite{kim2023coreset} with a $98.5\%$ reduction in sampling time with a mere $0.83\%$ average trade-off in accuracy calculated across $11$ downstream datasets. We have made the \href{https://github.com/prajwalsingh/BloomCoreset}{code} publicly available.
\end{abstract}


\section{Introduction}
\label{sec:intro}

The rise of deep neural networks (DNNs) has revolutionized various fields, including image recognition \cite{he2016deep}, semantic segmentation \cite{badrinarayanan2017segnet,long2015fully}, image synthesis \cite{chen2017photographic,rombach2022high}, and 3D analysis \cite{qi2017pointnet,fischer2021stickypillars}. These models often exceed human performance in tasks like computer vision, speech recognition, and natural language processing, thanks to extensive training on large labeled datasets. However, this reliance on vast annotated data poses a significant challenge, particularly in specialized domains like medical imaging, where expert annotations are scarce \cite{fotedar2020extreme, shurrab2022self, upadhyay2024advances}. Inspired by \cite{kim2023coreset}, our work introduces a sampling algorithm to construct a data subset (referred to as the Core-set) for domain-specific tasks such as classification using a vast unlabeled dataset referred to as the Open-Set. This approach is particularly valuable in contexts requiring differentiation between subtly distinct images, operating under limited labeled samples' constraints. 

\begin{table}[!t]
\caption{\textbf{Sampling Time.} We have compared the coreset sampling time between SimCore \cite{kim2023coreset} and our SimCore + BloomCoreset method. Our proposed sampling algorithm is order of magnitudes faster than the SimCore sampling algorithm. This is due to the pre-training of the self-supervised method required in the case of SimCore to sample the closest image from Open-Set. Here, the Open-Set is ImageNet1k \cite{deng2009imagenet}, and time is computed on two Nvidia RTX 3090 GPUs.}
\resizebox{\linewidth}{!}{
\def\arraystretch{1}
\begin{tabular}{lcccc}
\hline
         & \multicolumn{2}{c}{SimCore ~\protect\cite{kim2023coreset}}                                                                                                                       & \multicolumn{2}{c}{SimCore + BloomCoreset (ours)}                                                                                                                    \\ \cline{2-5} 
         & \begin{tabular}[c]{@{}c@{}}Preprocessing\\ (Sampling)\end{tabular} & \begin{tabular}[c]{@{}c@{}}Training \\ (Coreset+\\ Downstream )\end{tabular} & \begin{tabular}[c]{@{}c@{}}Preprocessing\\ (Sampling)\end{tabular} & \begin{tabular}[c]{@{}c@{}}Training\\ (Coreset+\\ Downstream)\end{tabular} \\ \hline
Open-Set & $\sim$23h 0m                                                       & $\sim$25h 0m                                                                 & \textbf{$\sim$0h 20m}                                                       & $\sim$25h 0m                                                               \\ \hline
\end{tabular}}
\label{table:time}
\vspace{-3mm}
\end{table}

\textbf{Self Supervised Learning:} Recent advancements in self-supervised learning (SSL) \cite{grill2020bootstrap,caron2021emerging} have paved the way for learning data representations without reliance on annotations, effectively improving the results of subsequent downstream tasks. Unlike unsupervised representation learning \cite{gidaris2018unsupervised,zhang2016colorful}, where the handcrafted pretext tasks are difficult to generalize and scale, contrastive learning \cite{chen2020simple,oord2018representation} focuses on creating positive pairs by image augmentation and using distinct images as negative samples. SSL benefits from the abundance of unlabeled data obtainable through web crawling \cite{goyal2021self}, thus removing the dependence on labeling, rendering it particularly advantageous. Our study focuses on the Open-Set Self-Supervised Learning (OpenSSL) task, as introduced by \cite{kim2023coreset}, where we utilize the large-scale Open-Sets for aiding the SSL pretraining using the specific fine-grained target datasets.

\textbf{Motivation for Coreset:} The work by Ericsson \emph{et al.} \cite{ericsson2021well} shows a correlation between the performance of self-supervised learned representations on downstream tasks and the similarity of the pretraining and fine-tuning datasets. This distribution mismatch issue is particularly pronounced in OpenSSL due to irrelevant instances in the Open-Set, which can impede learning useful representations. The work by \cite{kim2023coreset} tackles distribution mismatch by curating a coreset from the Open-Set that aligns semantically with limited fine-grained labeled samples, enhancing SSL efficacy on downstream tasks. This focused approach outperforms the case of using the entire Open-Set, as shown in Table \ref{table:simclrall}. 


\textbf{Coreset Sampling.} A coreset is a subset of an Open-Set that shares semantic similarities with the downstream dataset. The study by \cite{kim2023coreset} highlights the performance benefits of integrating a coreset with the downstream dataset $X$, compared to either the entire Open-Set $(X+OS)$ or a randomly sampled subset $(X+OS_{rand})$. However, existing sampling algorithms, such as those discussed in \cite{kim2023coreset,sener2017active,wei2015submodularity}, 
exhibit substantial \textit{computational latency}, rendering them impractical as off-the-shelf solutions and do not take into consideration the distribution discrepancies in the Open-Set. To address these gaps, our research proposes BloomCoreset as a novel strategy that drastically improves the coreset sampling speed while preserving competitive performance compared to the state-of-the-art baseline sampling method, as shown in Table\ref{table:time}. 

\begin{figure}[!t]
  \centering
  \includegraphics[width=1.0\linewidth]{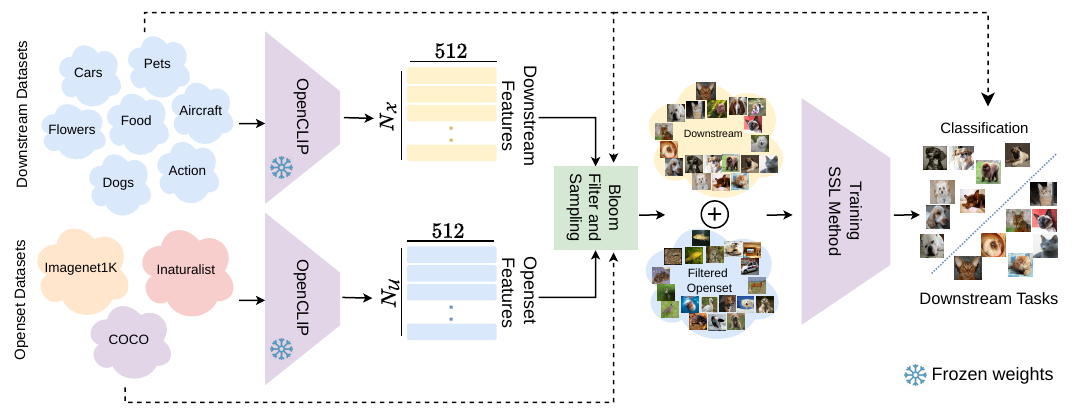}
  \caption{\textbf{Fine-Grained SSL Framework.} The figure illustrates the workflow for addressing a fine-grained SSL problem. We use the pre-trained OpenCLIP model \cite{cherti2023reproducible} to extract image features (hash codes) from domain-specific and open-set image pools. These features are then processed through the BloomCoreset algorithm, which generates a coreset from the open-set image pool. The coreset and domain-specific data are subsequently used to train the self-supervised method.}
  \label{fig:block_diagram}
  \vspace{-3mm}
\end{figure}

The primary contributions of this work are: \textbf{1) Faster Samples Retrieval} - We employ a probabilistic data structure Counting Bloom Filter/Bloom Filter \cite{fan2000summary,bloom1970space} which aids in significantly expediting the retrieval of semantically similar images from large Open-Sets as shown in Fig.\ref{fig:block_diagram} and Fig.\ref{fig:samplingdiag}, drastically outperforming the baseline with a 98.5\% reduction in sampling speed. \textbf{2) Robust Cross-Dataset Performance} - Our work demonstrates competitive performance over existing sampling strategies across a wide range of fine-grained datasets (Table \ref{table:simclrall}) and Open-Sets (Fig.\ref{fig:compalldataset}), as evidenced by extensive experimental evaluations with a mere 0.83\% average trade-off in accuracy across 11 downstream datasets.

\section{Method}
\label{sec:method}

In this work, we introduce a coreset sampling method that acts as a module for fine-grained self-supervised learning, as shown in Fig.\ref{fig:block_diagram}. The proposed modular framework consists of two phases: 1) Generating features for downstream dataset and Open-Set using pre-trained Open-CLIP \cite{cherti2023reproducible}. These generated features are used to build the coreset by employing a bloom filter. To tackle the false positive characteristics of probabilistic data structures, we employ top-k filtering, leveraging an adaptive similarity metric to threshold and prune the candidate set, thereby optimizing the trade-off between computational efficiency and sample representativeness. 2) Using the sampled coreset and downstream data to train a fine-grained SSL method that can be utilized for domain-specific tasks. 

By leveraging the probabilistic data structure and controlling the false positive characteristics, our algorithm improves the sampling time by orders of magnitude while maintaining the quality and fidelity of the selected samples.

\subsection{Problem Formulation}
\label{sec:background}

In alignment with the paradigm established by \cite{kim2023coreset}, our methodology entails the generation of augmented pairs $(x_{i}^{t_{1}}, x_{i}^{t_{2}})$ from the given data corpus $\mathcal{X}=\left\{x_{i}\right\}_{i=1}^{N}$. An encoder network $E_{\theta}$ is trained on a random batch from training data $(\mathcal{X_{B}})$ with the objective of the minimization of the contrastive loss as proposed in \cite{chen2020simple} by projecting the positive pair $(x_{i}^{t_{1}}, x_{i}^{t_{2}})$ close in the representation space and contrasting it from all the negative pairs $\{(x_{j}^{t_{1}}, x_{j}^{t_{2}}) | j \neq i, (i,j) \in \{1, 2, ..., N_{B}\} \}$.

\begin{align*}
    &\mathcal{L}_{c} = \underset{x_{i} \in \mathcal{X_{B}}, x_{j} \in \mathcal{X_{B}}}{\mathbb{E}} \Bigg[-log\frac{exp(sim(z_{i}, z_{j})/\tau)}{\sum\limits_{k=1}^{2N_{B}}\mathbbm{1}_{[k \neq i]} exp(sim(z_{i}, z_{k})/\tau)} \Bigg] \notag \\
\end{align*}

Here, $L_{c}$ is the contrastive loss, $z = E_{\theta}\left(x^{t}\right)$ denotes the representation from encoder $E$ parameterized over $\theta$ which are the learnable parameters. $N_{B}$ is the number of samples in the image batch and $(t_{1}, t_{2})$ are the pair of transformations from transformation space $\mathcal{T}$.

\textit{In the context of the OpenSSL problem, our objective is to sample such data from Open-Set ($\mathcal{U}$) whose inclusion in the downstream dataset ($\mathcal{X}$) improves the performance of representation learning of the fine-grained SSL method.}

\begin{figure}[!t]
\centering
 \includegraphics[width=1.0\linewidth]{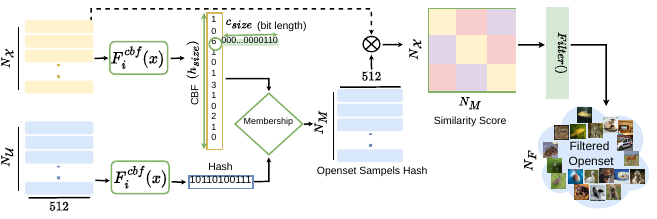}
\caption{\textbf{BloomCoreset.} Domain-specific features are used to build the Counting Bloom Filter (CBF). A membership test is then conducted to sample Open-Set images similar to domain-specific ones. After sampling, the inner product is calculated, and additional filtering is applied to select the best subset from the sampled Open-Set images. A detailed overview of the sampling method is provided in Section \ref{subsec:coresetexpalin}.}
\label{fig:samplingdiag}
\vspace{-3mm}
\end{figure}

\subsection{Coreset Sampling from Open-Set}
\label{subsec:coresetexpalin}

The work by \cite{kim2023coreset} initially trains a self-supervised model on the downstream dataset and then samples the closest images from an Open-Set. However, this approach of training and sampling is time-consuming and resource-intensive, particularly with the recent trend of using multi-modal SSL methods trained on extensive datasets containing millions of data points. Our work introduces a more efficient sampling algorithm in response to this challenge.

Recent advances have seen the use of billions of data points \cite{schuhmann2022laionb} to train multi-modal self-supervised models \cite{Radford2021LearningTV,cherti2023reproducible}, with pre-trained weights readily available for fine-tuning or testing. We leverage such a pre-trained Open-CLIP model \cite{cherti2023reproducible} as the feature extractor for both downstream and Open-Set data, as shown in Fig.\ref{fig:block_diagram}.

\begin{figure}[!t]
\begin{minipage}{\linewidth}
\removelatexerror
\begin{algorithm}[H]
\DontPrintSemicolon  
  \KwInput{$E_{\theta}$: pre-trained open clip image encoder}
  \KwInput{$\mathcal{U}$: all the images in open-set}
  \KwInput{$\mathcal{X}$: all the images in downstream dataset}
  \KwInput{$CBloomFilter()$: method that returns object of CBF, $\mathcal{B}$: coreset budget}
  Initialize $N_{\mathcal{X}} \xleftarrow{} |\mathcal{X}|$, $h_{size} \xleftarrow{} 10000 \times (N_{\mathcal{X}}/3500) $ \\
   Initialize $c_{size} \xleftarrow{} 32 $, $feat_{\mathcal{X}} \xleftarrow{} \emptyset$\\
  Initialize $feat_{\mathcal{U}} \xleftarrow{}  \emptyset$, $member_{\mathcal{U}} \xleftarrow{}  \emptyset$ \\
  $cbloom \xleftarrow{} CBloomFilter(h_{size}, c_{size})$ \\
  \tcc{Creating counting bloom filter (CBF)}
  \For{$I_{\mathcal{X}} \in \mathcal{\mathcal{X}}$}
   {
        $z \xleftarrow{} E_{\theta}(I_{\mathcal{X}})$ \tcp*{ $z \in \mathbb{R}^{512}$}
        $feat_{\mathcal{X}}.append(z)$ \\
        $bin_{z} \xleftarrow{} where(z < 0, 0, 1)$ \\
        $cbloom.update(bin_{z})$ \\
   }
   \tcc{Sampling from the open-set using CBF}
   \For{$I_{\mathcal{U}} \in \mathcal{U}$}
   {
        $z \xleftarrow{} E_{\theta}(I_{\mathcal{U}})$ \tcp*{ $z \in \mathbb{R}^{512}$}
        $feat_{\mathcal{U}}.append(z)$ \\
        $bin_{z} \xleftarrow{} where(z < 0, 0, 1)$ \\
        \If{$cbloom.check(bin_{z})$}
        {
            $member_{\mathcal{U}}.append(z)$
        }
   }
   \tcc{Refining the images from $member_{\mathcal{U}}$}
   $N_{M} \xleftarrow{} |member_{\mathcal{U}}|$ \\
   $simscore \xleftarrow{} feat_{\mathcal{X}} \times member_{\mathcal{U}}.T$ \tcp*{$\mathbb{R}^{N_{\mathcal{X}} \times N_{M}}$}
   $coreset \xleftarrow{} Filter(simscore, \mathcal{\mathcal{U}}, \mathcal{B})$ \\
   \KwOutput{$coreset$}
\caption{Sampling image subset (coreset) from the open-set images}
\label{algo:coresetsampling}
\end{algorithm}
\end{minipage}
\vspace{-5mm}
\end{figure}

To sample images closer in the feature space of the downstream dataset from the Open-Set, we introduce a framework utilizing Bloom Filters \cite{bloom1970space} as a hashing mechanism. Bloom Filters are probabilistic data structures optimized for membership testing with zero false negatives. We use Open-CLIP features from the downstream dataset to construct a Bloom Filter, employing the murmurhash3 \cite{murmurhash3} hash function $(F^{cbf})$ to build the hash table. Ten variations of murmurhash3 $(F_{i}^{cbf})$ are used to create the Bloom Filter, which then performs membership testing on the Open-CLIP features extracted from Open-Set images.

The time complexity for adding an item or checking membership in a Bloom Filter is $O(h)$, where $h$ is the number of hash functions used. This ensures that these operations remain constant in time, regardless of the data size. Membership testing requires only a single access to the data structure, significantly reducing memory access time. Thus, leveraging a pre-trained model trained on vast datasets and an efficient data structure like the Bloom Filter enables robust feature representation in a fraction of the time taken by \cite{kim2023coreset}.


\begin{figure}[!t]
    \begin{minipage}[b]{\linewidth}
        \centering
        \includegraphics[width=\linewidth]{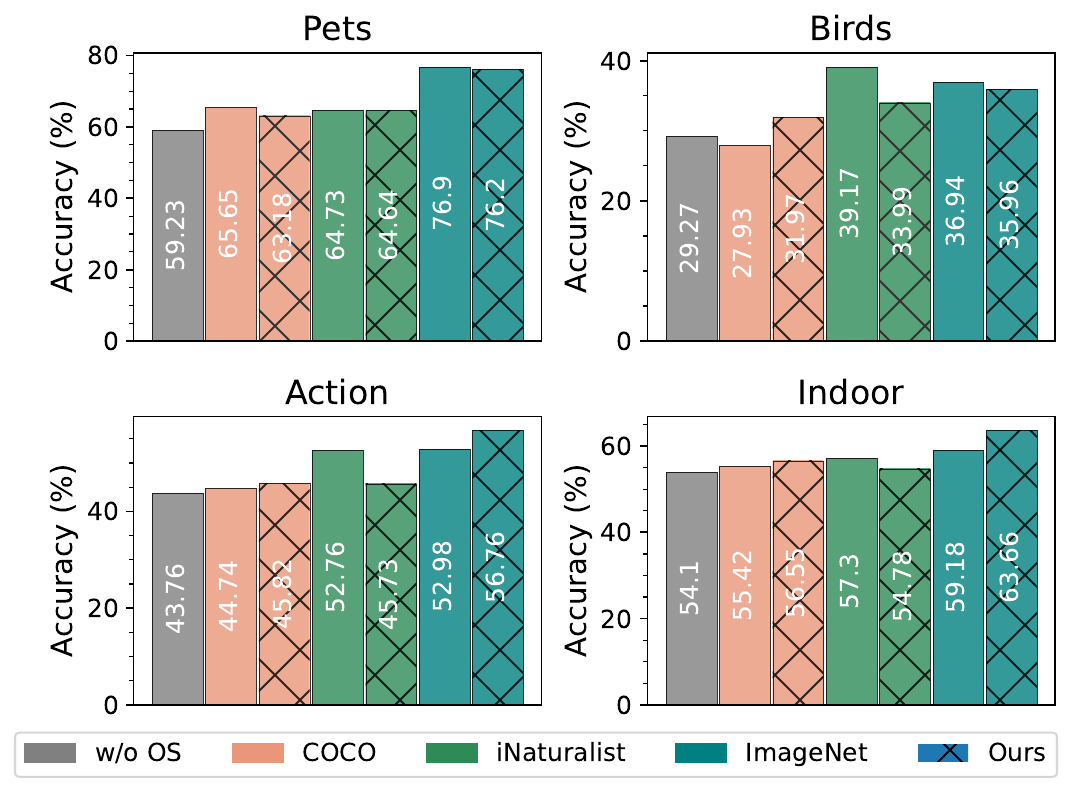}
        \captionof{figure}{\textbf{Other Open-Sets.} While comparing across different Open-Sets, the learned representation from the coreset sampled using the proposed method shows competitive performance to the baseline and, in some cases, outperforms the baseline.}
        \label{fig:compalldataset}
    \end{minipage}
    \vspace{-6mm}
\end{figure}

However, the coreset generated by Bloom Filters and the downstream dataset doesn't significantly enhance representation learning, as shown in Table \ref{table:randombloom}. This is due to false positives in the coreset, a known characteristic of Bloom Filters. To address this, we introduce a top-$k$ filtering method. In this approach, we compute cosine similarity scores between the filtered and downstream samples to retrieve Open-Set samples most similar to the downstream dataset (see Algorithm \ref{algo:coresetsampling}). The entire sampling pipeline is depicted in Fig.\ref{fig:samplingdiag}, where $N_{\mathcal{X}}$ represents the number of downstream samples, $N_{\mathcal{U}}$ the number of Open-Set samples, $N_{M}$ the filtered samples after membership testing, and $N_{F}$ is the final coreset size after applying top-$k$ filtering.

\begin{table*}[!t]
\centering
\caption{Comparison of classification accuracy between the representation learned using the coreset sampled by our proposed method and the baseline \cite{kim2023coreset}. The coreset generated by our method demonstrates competitive performance across many downstream datasets, outperforming the baseline in some cases. The last row highlights the minimal accuracy trade-off, with our method surpassing the baseline in some datasets.}
\resizebox{\linewidth}{!}{
\def\arraystretch{1.3}%
\begin{tabular}{lcccccccccccc} \hline
\multicolumn{1}{c}{}  &     & \multicolumn{11}{c}{Target dataset (X) and its number of samples}                               \\
\multicolumn{1}{c}{}  &     & Aircraft & Cars  & Pet   & Birds & Dogs  & Flowers & Action & Indoor & Textures & Faces & Food  \\ 
Pretrain              & p  & $6667$     & $8144$  & $3680$  & $5990$  & $12000$ & $2040$    & $4000$   & $5360$   & $3760$     & $4263$  & $13296$ \\ \hline
$\mathcal{X}$                     & -   & $46.56$    & $55.42$ & $59.23$ & $29.27$ & $49.88$ & $80.14$   & $43.76$  & $54.10$  & $58.78$    & $56.63$ & $87.99$ \\
$\mathcal{X}+OS$                  & -   & $39.88$    & $42.92$ & $68.22$ & $32.88$ & $50.42$ & $\underline{85.34}$   & $\textbf{60.61}$  & $\underline{63.66}$  & $\textbf{67.98}$    & $52.76$ & $85.90$ \\ 
$\mathcal{X}+OS_{rand}$         & 1\% & $48.24$    & $49.26$ & $64.27$ & $31.90$ &$ 49.62$ & $83.17$   & $47.25$  & $55.37$  & $61.33$    & $\underline{57.37}$ & $88.08$ \\ \hline
$\mathcal{X}+OS_{SimCore}$  \cite{kim2023coreset}    & 1\% & $\textbf{50.79}$    & $\textbf{57.90}$ & $\textbf{76.90}$ & $\textbf{36.94}$ & $\textbf{59.83}$ & $\textbf{86.70}$   & $52.98$  & $59.18$  & $63.40$    & $\textbf{58.85}$ & $\underline{89.78}$ \\
$\mathcal{X}+OS_{BloomCoreset}$ (\textbf{Ours}) & 1\% &    $\underline{49.34}$      &   $\underline{52.08}$    &   $\underline{76.20}$    &  $\underline{35.96}$    &   $\underline{54.43}$    &    $84.56$     &   $\underline{56.76}$     &    $\textbf{63.66}$    &     $\underline{65.11}$     &   $55.72$    &    $\textbf{90.20}$   \\ \hline
Acc. Trade-off (\textbf{Ours} - \cite{kim2023coreset}) & - &    \textcolor{red}{$-1.45\%$} &   \textcolor{red}{$-5.1\%$}    &   \textcolor{red}{$-0.7\%$}    &  \textcolor{red}{$-0.98\%$}    &   \textcolor{red}{$-5.4\%$}    &    \textcolor{red}{$-2.14\%$}     &   \textcolor{blue}{$+3.78\%$}     &    \textcolor{blue}{$+4.48\%$}    &     \textcolor{blue}{$+1.71\%$}     &   \textcolor{red}{$-3.13\%$}    &    \textcolor{blue}{$+0.42\%$}   \\ \hline

\end{tabular}}
\label{table:simclrall}
\vspace{-3mm}
\end{table*}
\section{Experiments}
\label{sec:experiments}


\subsection{Network Training and Datasets}

We have used the eleven downstream datasets and three Open-Sets for evaluation and comparison: Aircraft \cite{maji2013fine}, Cars \cite{krause20133d}, Pet \cite{parkhi2012cats}, Birds \cite{welinder2010caltech}, Dogs \cite{khosla2011novel}, Flowers \cite{nilsback2008automated}, Action \cite{yao2011human}, Indoor \cite{quattoni2009recognizing}, Textures \cite{cimpoi2014describing}, Faces \cite{lee2020maskgan}, Food \cite{singla2016food}, ImageNet1k \cite{deng2009imagenet}, MS COCO \cite{lin2014microsoft}, and iNaturalist 2021-mini \cite{van2018inaturalist}.

For the self-supervised learning backbone, we used SimCLR \cite{chen2020simple} with ResNet50. We have trained the network on two GPUs with a batch size of $256$. For a fair comparison, we have kept all the experimental setups similar to the SimCore \cite{kim2023coreset} and re-run most of the experiments of SimCore with batch size $256$ on two GPUs. In all the experiments, including SimCore and ours, we kept a threshold on the size of the coresets as $1\%$, i.e., we consider only $1\%$ of OpenSet data for training the SSL method.

\subsection{Analysis of Sampling Method}

Using Bloom Filters, we sampled each downstream dataset's coreset from ImageNet1k \cite{deng2009imagenet}. The sampled coreset and downstream data were then used to train the SimCLR \cite{chen2020simple} network. We followed the linear evaluation strategy for comparison with the baseline, employing the pre-trained SimCLR network as a feature extractor and training an image classification model in a supervised setting.


We employed a variation of the Bloom Filter known as the Counting Bloom Filter \cite{5374398}, which tracks the frequency of hash location accesses during data storage. This feature was utilized to record the frequency of samples in both the downstream dataset and the sampled coreset. The coreset sampling process is outlined in Fig.\ref{fig:samplingdiag}.

\begin{table}[t]
\centering
\caption{\textbf{Random Bloom.} Quantitative results - Using samples from a bloom filter without applying the proposed post-filtering approach leads to degradation in performance.}
\begin{tabular}{lccccc}\hline
         &     & \multicolumn{4}{c}{Target Dataset (X)} \\
         &     & Pet                & Flowers              & Textures    & Action         \\
Pretrain & p   & $3680$               & $2040$                 & $3760$       & 4000          \\ \hline
$\mathcal{X}$ + $OS_{BloomRandom}$   & $1\%$ & $67.56$              & $82.16$                & $62.66$   &  $50.89$    \\
$\mathcal{X}$ + $OS_{BloomCoreset}$   & $1\%$ & $\textbf{76.20}$              & $\textbf{84.56}$                & $\textbf{65.11}$        &  $\textbf{56.76}$      \\
\hline
\end{tabular}
\label{table:randombloom}
\vspace{-3mm}
\end{table}

\textbf{Sampling and Classification.} Table \ref{table:simclrall} compares the coreset quality between the SimCore method \cite{kim2023coreset} and our BloomSSL method. Our approach demonstrates competitive performance in representation learning, with some downstream datasets showing better results than the baseline. This highlights the trade-off between coreset sampling speed and quality. While conventional methods may yield fewer false positives, they are significantly more time-consuming. In contrast, BloomSSL offers faster sampling with only a minimal accuracy trade-off which is controlled using top-$k$ filtering. Table \ref{table:time} compares the time required for BloomSSL and SimCore \cite{kim2023coreset}, and Table \ref{table:randombloom} shows that using Bloom Filter samples without top-$k$ filtering results in degraded performance.

\begin{figure}[!t]
    \begin{minipage}[b]{\linewidth}
        \centering
        \resizebox{\linewidth}{!}{\begin{tabular}{ccccc}
            & \LARGE{Clip Downstream} & \LARGE{Clip Coreset} & \LARGE{Pre-trained Downstream} & \LARGE{Pre-trained Coreset} \\
           \rotatebox[origin=c]{90}{\huge{(a) Aircraft}} & \includegraphics[width=0.69\linewidth,valign=m]{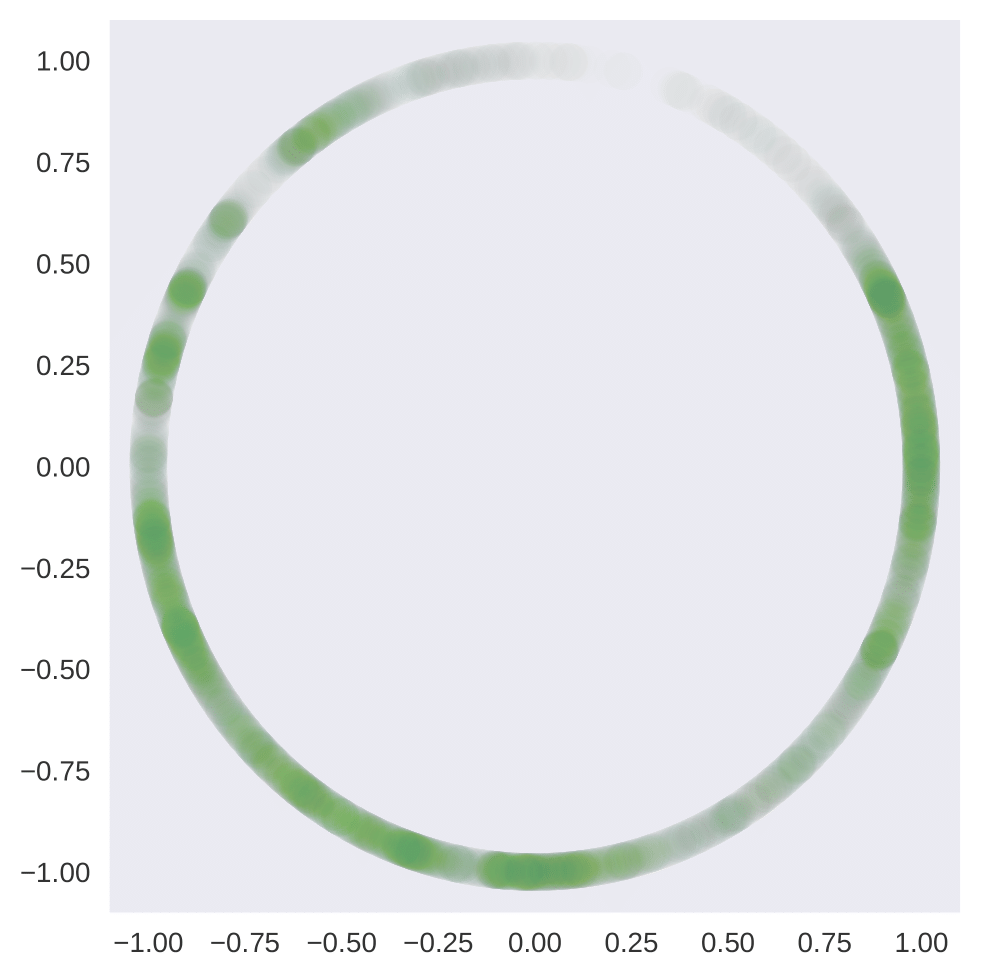}   &  \includegraphics[width=0.69\linewidth,valign=m]{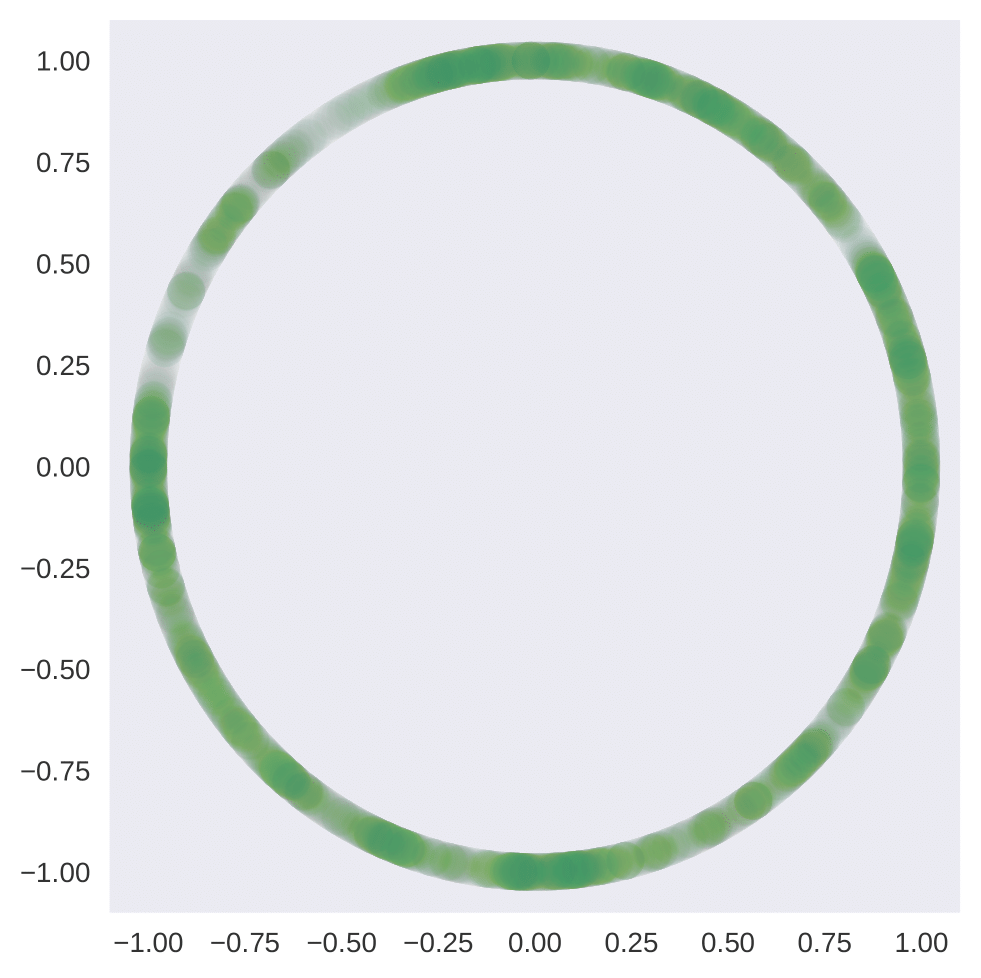}   &  \includegraphics[width=0.69\linewidth,valign=m]{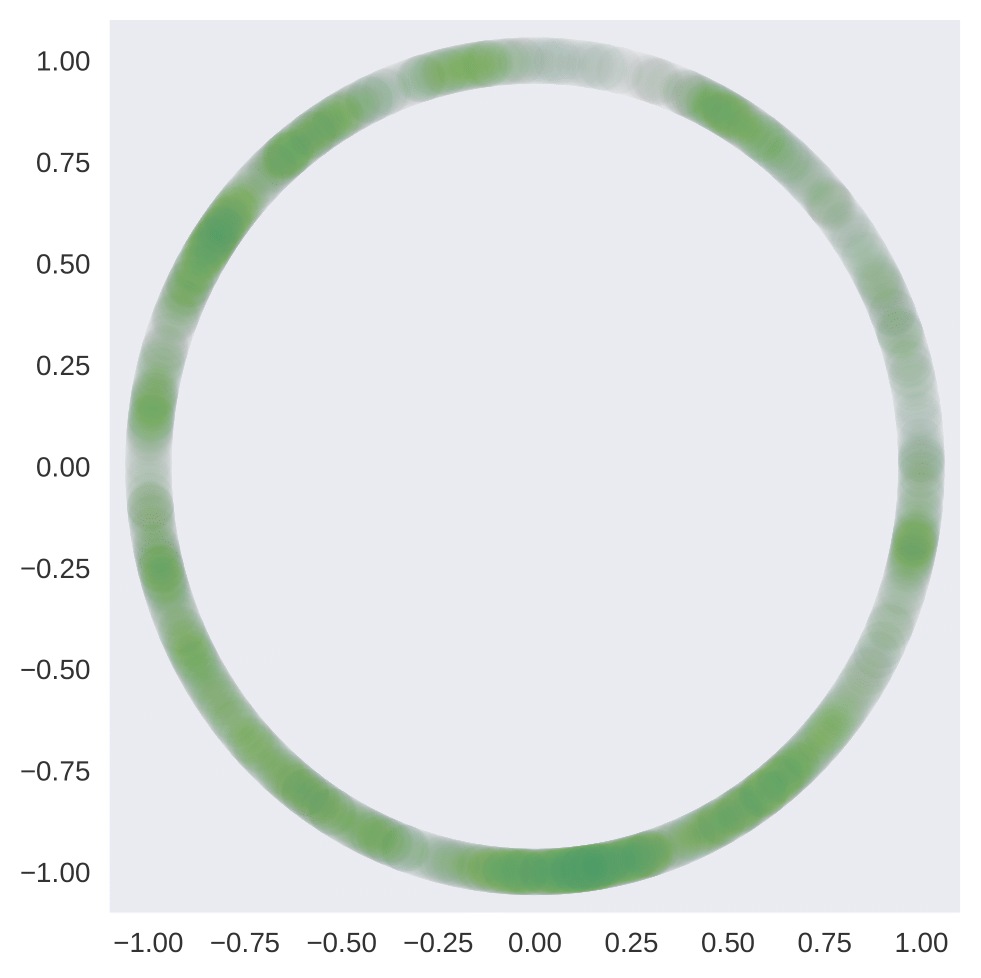}   &  \includegraphics[width=0.69\linewidth,valign=m]{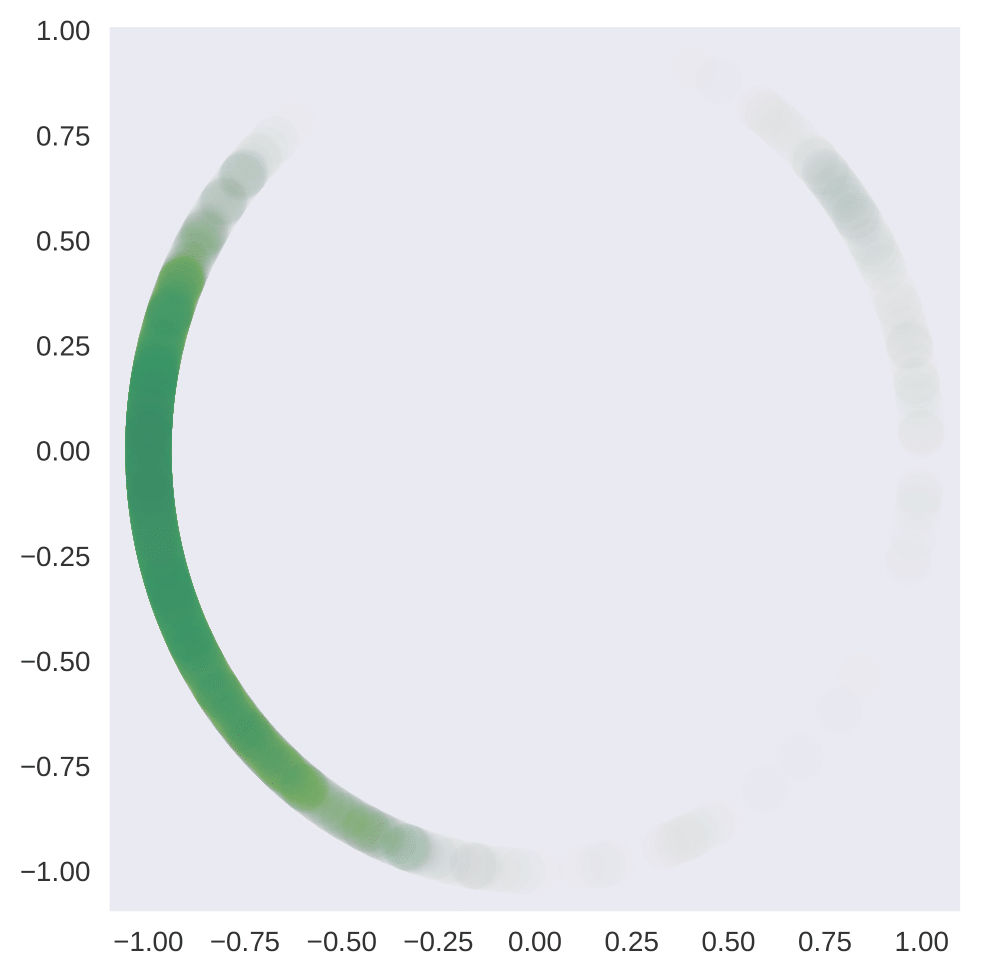}      \\
            \rotatebox[origin=c]{90}{\huge{(b) Pets}} & \includegraphics[width=0.69\linewidth,valign=m]{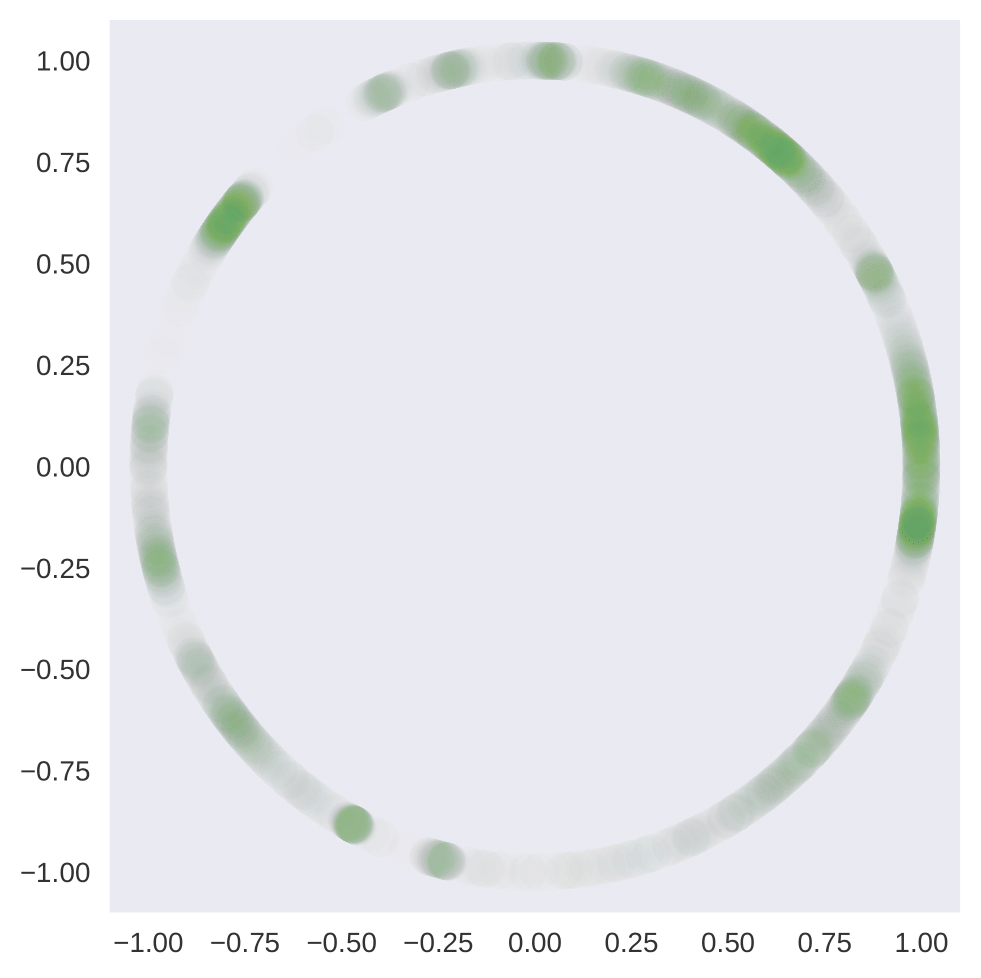}   &  \includegraphics[width=0.69\linewidth,valign=m]{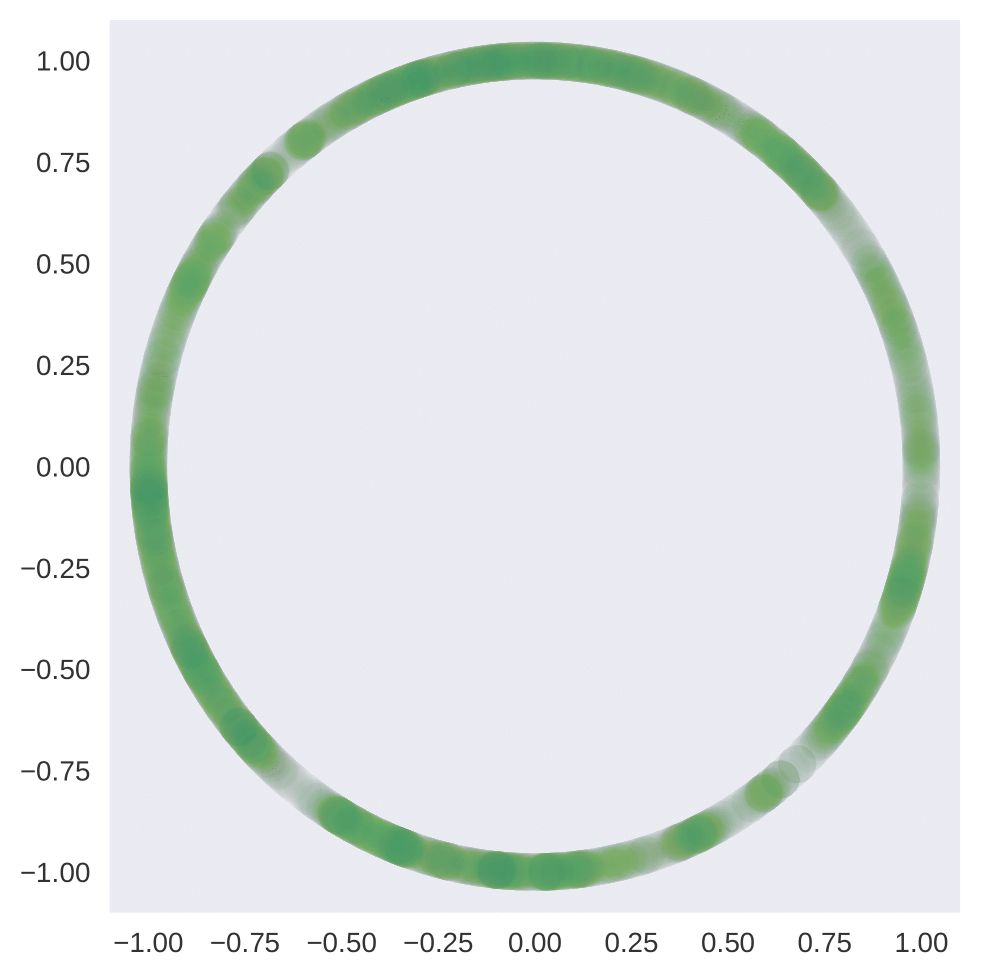}   &  \includegraphics[width=0.69\linewidth,valign=m]{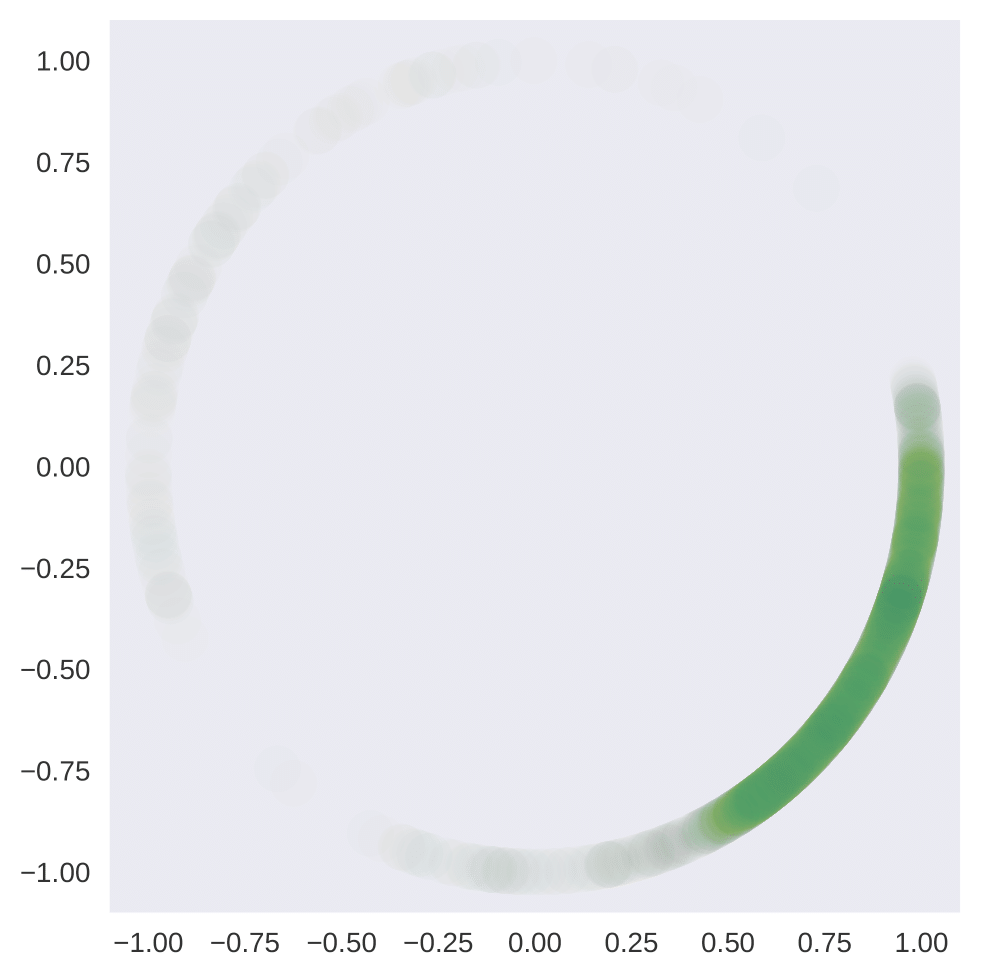}   &  \includegraphics[width=0.69\linewidth,valign=m]{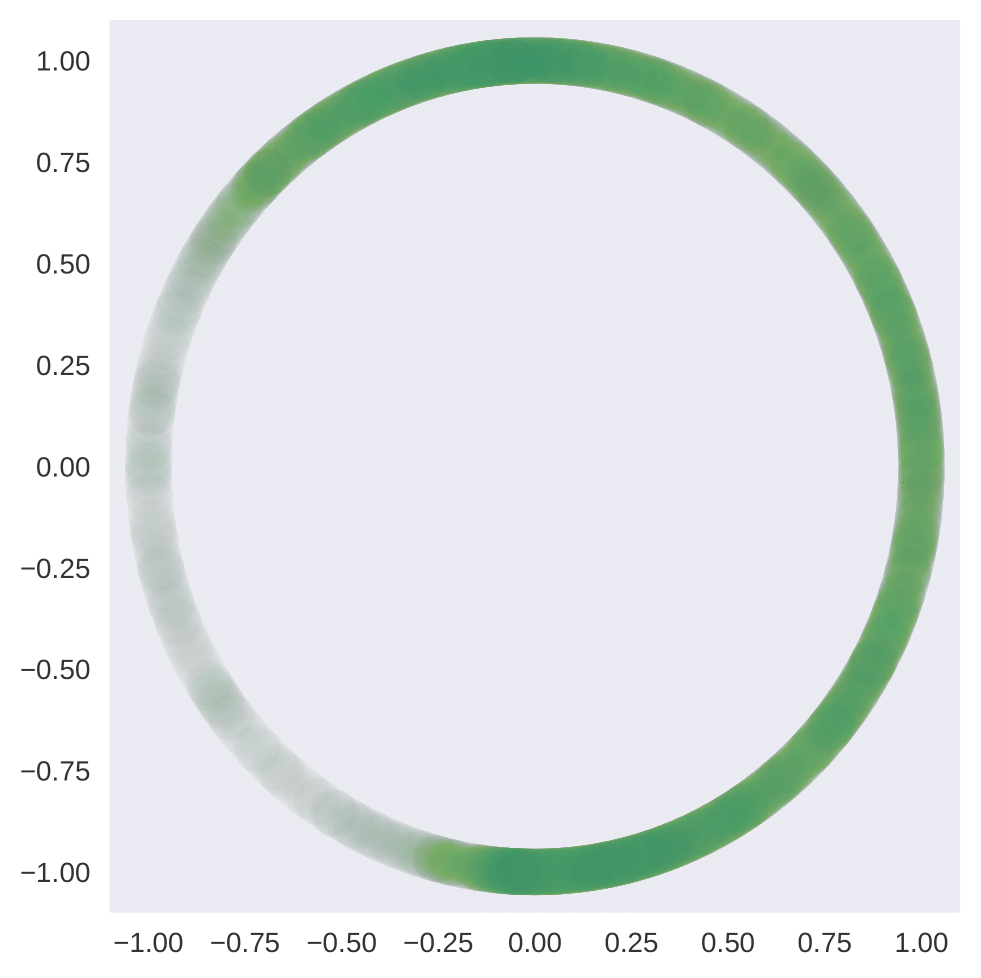}      \\
            \rotatebox[origin=c]{90}{\huge{(c) Food}} & \includegraphics[width=0.69\linewidth,valign=m]{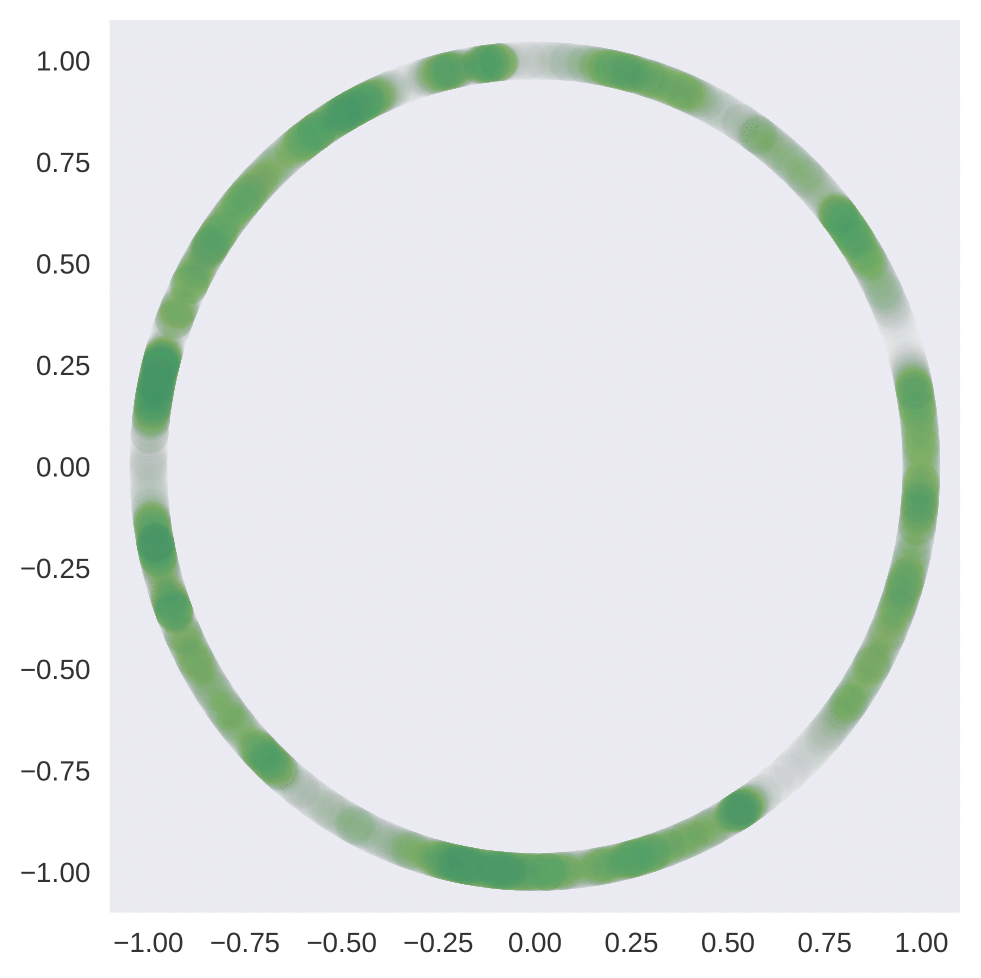}   &  \includegraphics[width=0.69\linewidth,valign=m]{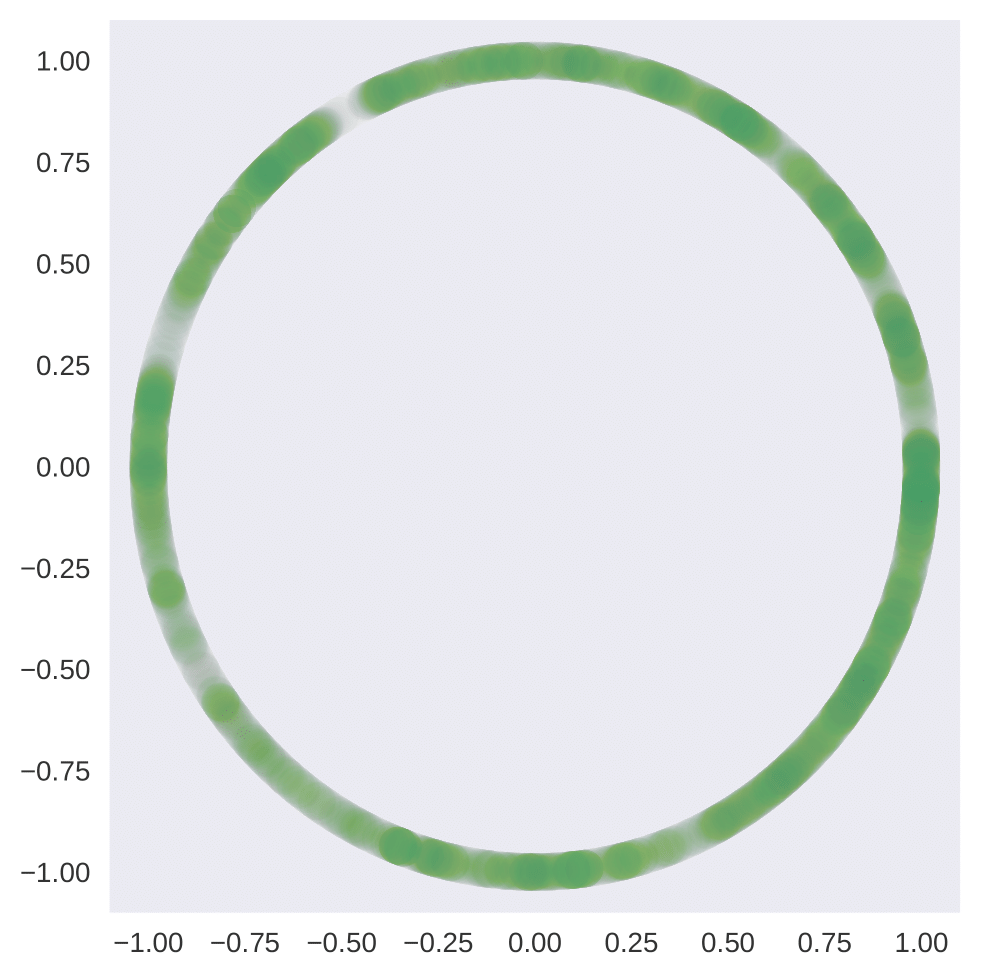}   &  \includegraphics[width=0.69\linewidth,valign=m]{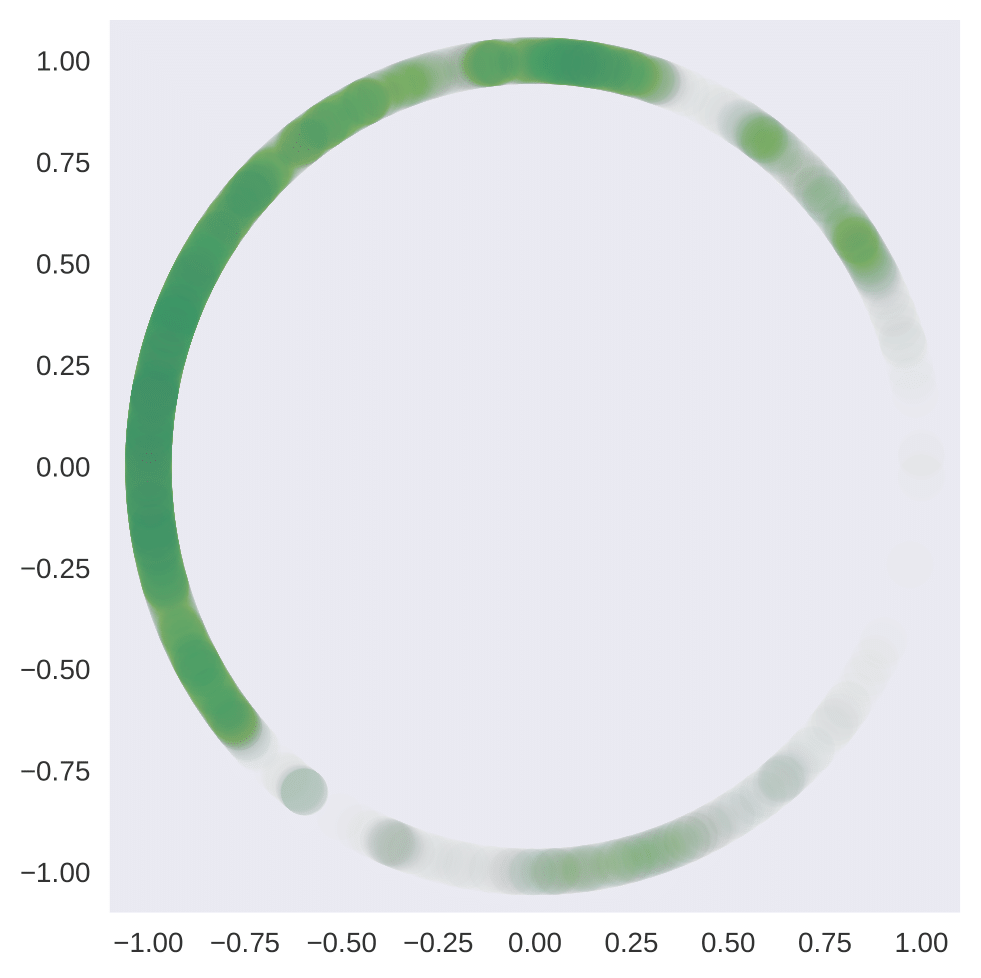}   &  \includegraphics[width=0.69\linewidth,valign=m]{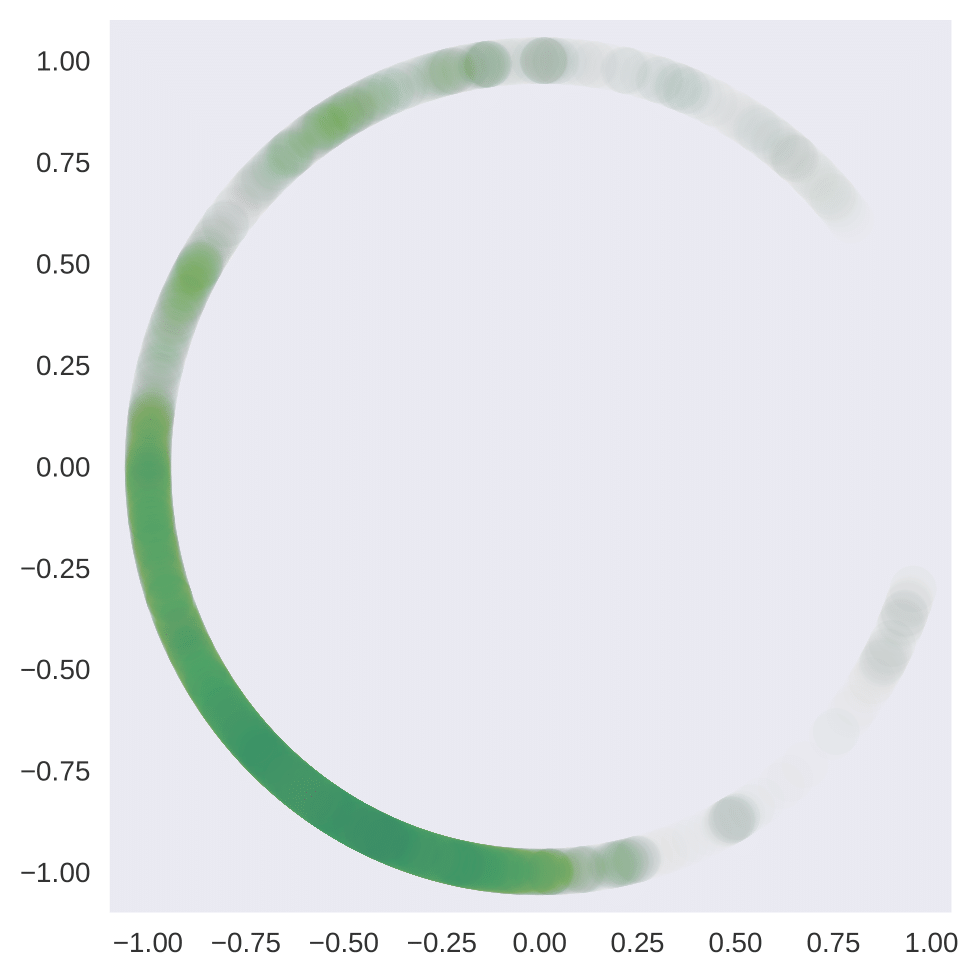}
        \end{tabular}}
        \captionof{figure}{\textbf{Feature Distribution.} Show representation space of Open-CLIP ~\protect\cite{cherti2023reproducible}. We have used Gaussian kernel density estimation ~\protect\cite{wang2020understanding} to show the feature distribution of downstream dataset samples and coreset samples across the unit ring.}
        \label{fig:featdist}
    \end{minipage}
\end{figure}

\textbf{Various Open-Sets.} We tested the BloomSSL method on a range of Open-Sets beyond ImageNet1k, including MS COCO \cite{lin2014microsoft} and iNaturalist 2021-mini \cite{van2018inaturalist}. Fig.\ref{fig:compalldataset} illustrates performance improvements across various downstream datasets using different Open-Sets, compared to the SimCore \cite{kim2023coreset} method, shows the generalizability of our coreset sampling approach.

\textbf{Feature Density Map.} The density mapping in Fig.\ref{fig:featdist} shows that the pre-trained Open-CLIP model generates semantically rich features that cover the entire unit ring \cite{wang2020understanding}, thanks to its multi-modal training. Without Bloom Filter and top-$k$ filtering, the coreset would include diverse Open-Set images, which could degrade performance. The pre-trained downstream plot in Fig.\ref{fig:featdist} compares the distribution ring when the SSL method is pre-trained with only downstream data versus when pre-trained with coreset data. This comparison demonstrates that the data sampled from Open-Set using BloomCoreset aligns well with the downstream dataset distribution.

\section{Discussion}
\label{sec:discussion}

Table \ref{table:time} demonstrates that our proposed algorithm significantly reduces sampling time by \textbf{98.5\%} compared to \cite{kim2023coreset}, with only a \textbf{0.83\%} average tradeoff in accuracy across $11$ datasets. Our method outperforms the state-of-the-art in $4$ datasets but does not consistently surpass all datasets. Future work could refine the sampling algorithm to select semantically richer coresets for improved performance across more datasets.
\section{Conclusion}
\label{sec:conclusion}

This paper addressed the problem of coreset sampling for the OpenSSL task. The proposed work primarily focused on improving the coreset sampling time while maintaining the accuracy of downstream tasks for off-the-shelf use cases. With several empirical studies, we demonstrate the effectiveness of the proposed BloomCorest method in a fine-grained SSL setting. In future work, we aim to explore advanced strategies that maintain efficiency gains and bridge the performance gap across all datasets, potentially setting new benchmarks for speed and accuracy in SSL frameworks.

\section{Acknowledgment}
This work is supported by the Prime Minister Research Fellowship (PMRF2122-2557) awarded to Prajwal Singh and by the Jibaben Patel Chair in Artificial Intelligence held by Shanmuganathan Raman.

\bibliographystyle{IEEEtran}
\small{
\bibliography{IEEEfull}
}
\end{document}